\documentclass{article} %
\usepackage{iclr2024_conference,times}
\pdfoutput=1

\usepackage{amsmath,amsfonts,bm}

\def\eqref#1{equation~\ref{#1}}

\def\1{\bm{1}}

\def\vp{{\bm{p}}}

\DeclareMathAlphabet{\mathsfit}{\encodingdefault}{\sfdefault}{m}{sl}
\SetMathAlphabet{\mathsfit}{bold}{\encodingdefault}{\sfdefault}{bx}{n}

\newcommand{\softmax}{\mathrm{softmax}}

\newcommand{\KL}{D_{\mathrm{KL}}}

\DeclareMathOperator*{\argmax}{arg\,max}

\usepackage{hyperref}
\usepackage{url}
\usepackage{booktabs}
\usepackage{multirow}
\usepackage{multicol}
\usepackage{graphicx}
\usepackage{wrapfig}
\usepackage{xspace}
\usepackage{amsmath,amssymb}
\usepackage{makecell}
\usepackage{diagbox}
\usepackage{caption}
\usepackage{subcaption}
\usepackage{siunitx}
\graphicspath{ {figures/} }
\title{On the Learnability of Watermarks \\for Language Models}

\author{Chenchen Gu, Xiang Lisa Li, Percy Liang, Tatsunori Hashimoto \\
Stanford University \\
\texttt{\{cygu, xlisali, thashim\}@stanford.edu, pliang@cs.stanford.edu}
}

\newcommand{\Vocab}{\mathcal{V}}
\newcommand{\fw}{f_\text{w}}
\newcommand{\pw}{p_\text{w}}
\newcommand{\fd}{f_\text{d}}
\newcommand{\fhash}{f_\text{hash}}
\newcommand{\KGW}{\text{KGW}\xspace}
\newcommand{\Aar}{\text{Aar}\xspace}
\newcommand{\KTH}{\text{KTH}\xspace}
\newcommand{\decbased}{decoding-based\xspace}
\newcommand{\Decbased}{Decoding-based\xspace}
\newcommand{\logitbased}{logit-based\xspace}
\newcommand{\Logitbased}{Logit-based\xspace}
\newcommand{\sampbased}{sampling-based\xspace}
\newcommand{\Sampbased}{Sampling-based\xspace}
\newcommand{\LlamaTwoSevenB}{Llama 2 7B\xspace}
\newcommand{\LlamaTwoChatSevenB}{Llama 2-Chat 7B\xspace}

\newcommand{\DD}{\mathcal{D}}
\newcommand{\Loss}{\mathcal{L}}
\newcommand{\llogitbased}{\Loss_\mathrm{logit}}
\newcommand{\lsampbased}{\Loss_\mathrm{sampling}}
\newcommand{\plm}{p_{\mathrm{LM}}}
\newcommand{\ptheta}{p_\theta}

\newcommand{\wbased}{weights-based\xspace}

\newcommand{\teststatcolor}[1]{\textcolor{gray}{#1}}
\newcommand{\tsc}[1]{\teststatcolor{(#1)}}

\newcommand{\len}{\mathrm{len}}
\newcommand{\onehot}{\mathrm{onehot}}
\newcommand{\Bin}{\mathrm{Bin}}

\newcommand{\RotateTableHeader}[1]{\rotatebox[origin=l]{90}{#1}}
\newcommand{\DecodingTable}{\RotateTableHeader{Decoding}}
\newcommand{\LogitTable}{\RotateTableHeader{Logit}}
\newcommand{\SamplingTable}{\RotateTableHeader{Sampling}}
\newcommand{\LlamaTableHeader}{
    \toprule
    \multicolumn{2}{c|}{} & \multicolumn{3}{c|}{\makecell{\textbf{p-value} $\left(\downarrow\right)$ \\ \tsc{\KTH test statistic $\left(\downarrow\right)$}}} & \multicolumn{3}{c|}{\textbf{AUROC} $\left(\uparrow\right)$} & \multicolumn{3}{c|}{\textbf{Perplexity} $\left(\downarrow\right)$} & \multicolumn{3}{c}{\textbf{seq-rep-3} $\left(\downarrow\right)$} \\
    \multicolumn{2}{c|}{\textbf{Watermark}} & \DecodingTable & \LogitTable & \SamplingTable & \DecodingTable & \LogitTable & \SamplingTable & \DecodingTable & \LogitTable & \SamplingTable & \DecodingTable & \LogitTable & \SamplingTable \\
    \midrule
}
\newcommand{\RotatePythiaTableHeader}[1]{\rotatebox[origin=l]{90}{#1}}
\newcommand{\DecodingTeacherTable}{\RotatePythiaTableHeader{Decoding (T)}}
\newcommand{\DecodingStudentTable}{\RotatePythiaTableHeader{Decoding (S)}}
\newcommand{\SamplingPythiaTable}{\RotatePythiaTableHeader{Sampling}}
\newcommand{\PythiaTableHeader}{
    \toprule
    \multicolumn{2}{c|}{} & \multicolumn{2}{c|}{\makecell{\textbf{p-value} $\left(\downarrow\right)$ \\ \tsc{test stat. $\left(\downarrow\right)$}}} & \multicolumn{2}{c|}{\textbf{AUROC} $\left(\uparrow\right)$} & \multicolumn{3}{c|}{\textbf{Perplexity} $\left(\downarrow\right)$} & \multicolumn{3}{c}{\textbf{seq-rep-3} $\left(\downarrow\right)$} \\
    \multicolumn{2}{c|}{\textbf{Watermark}} & \DecodingTeacherTable & \SamplingPythiaTable & \DecodingTeacherTable & \SamplingPythiaTable & \DecodingTeacherTable & \DecodingStudentTable & \SamplingPythiaTable & \DecodingTeacherTable & \DecodingStudentTable & \SamplingPythiaTable \\ 
    \midrule
}

\iclrfinalcopy %
\begin{document}

\maketitle

\begin{abstract}
Watermarking of language model outputs enables statistical detection of model-generated text, which can mitigate harms and misuses of language models. Existing watermarking strategies operate by altering the decoder of an existing language model. In this paper, we ask whether language models can directly \emph{learn} to generate watermarked text, which would have significant implications for the real-world deployment of watermarks. First, learned watermarks could be used to build open models that naturally generate watermarked text, enabling watermarking for open models, where users can control the decoding procedure. Second, if watermarking is used to determine the provenance of generated text, an adversary can hurt the reputation of a victim model by spoofing its watermark and generating damaging watermarked text. To investigate the learnability of watermarks, we propose watermark distillation, which trains a student model to behave like a teacher model that uses decoding-based watermarking. We test our approach on three decoding-based watermarking strategies and various hyperparameter settings, finding that models can learn to generate watermarked text with high detectability. We also find limitations to learnability, including the loss of watermarking capabilities under fine-tuning on normal text and high sample complexity when learning low-distortion watermarks.\footnote{See \url{https://github.com/chenchenygu/watermark-learnability} for code and models.}
\end{abstract}

\section{Introduction}

As language models (LMs) become more capable and widely used, watermarking LM outputs becomes increasingly important to mitigate potential harms and misuses of LMs. Watermarking enables statistical detection of LM-generated text, which enables enforcing policies on LM usage, e.g., removing LM-generated disinformation from social media platforms or detecting academic dishonesty. Another proposed use case of watermarking is identifying the provenance of text, i.e., tracing text to the specific LM that generated it \citep{abdelnabi2021adversarial,kuditipudi2023robust}.

Recent works have shown that it is possible for an LM provider to inject specific, known watermark signals into text using specialized decoding algorithms \citep{kirchenbauer2023, aaronson2023, kuditipudi2023robust}, but little is known about whether these watermarks are learnable by a model. The learnability of watermarks has significant implications for the real-world deployment of watermarks, as it could enable downstream applications and adversarial spoofing attacks.

In this work, we study the learnability of watermarks by studying \textit{\wbased watermarking}, which involves learning parameters for a language model that cause it to generate watermarked text under its natural sampling distribution, without using a special decoding-time watermarking algorithm. Our investigation is motivated by its relevant implications for two applications: (i) developing watermarking for open language models and (ii) spoofing watermarks.

First, existing watermarking methods depend upon using a specialized decoding algorithm, making them too inflexible for open LMs. For open LMs, where the weights are released, a user can use an ordinary decoding algorithm and generate non-watermarked text, whether intentionally or not. We find that \wbased watermarking works with standard decoding strategies, removing the reliance on a specialized decoder. This makes it a promising first step towards developing watermarking for open LMs. However, we also find that \wbased watermarking capabilities can be removed by fine-tuning on normal text, indicating that improving robustness to fine-tuning is an important remaining challenge.

Second, in watermark spoofing attacks, an adversary outputs text that contains the watermark signal from a victim LM \citep{sadasivan2023aigenerated}. If watermarking is used to identify the provenance of text, then an attacker could attribute damaging text to the victim LM and hurt its reputation. We find that the learning of \wbased watermarking can enable spoofing attacks, and we demonstrate a proof-of-concept attack on an instruction-following chat model. The possibility of spoofing attacks suggests that watermarking should not be used to attribute provenance or blame to a specific LM. Instead, watermarking should only be used to statistically detect LM-generated text, which can be used for tasks such as finding infractions of policies on LM usage.

\begin{figure}
    \centering
    \includegraphics[width=\textwidth, page=1]{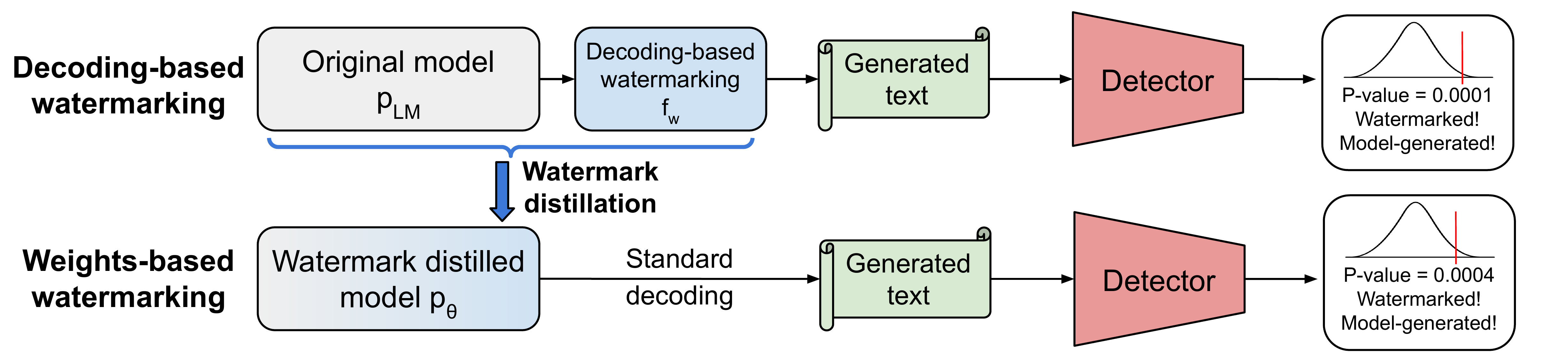} 
    \caption{\Decbased watermarking (top) versus \wbased watermarking (bottom). \Decbased watermarking requires a specialized decoding algorithm $\fw$ to generate watermarked text, whereas \wbased watermarking can use standard decoding to generate watermarked text directly from the model, using just its weights. Watermark distillation enables \wbased watermarking by training a student model $\ptheta$ to behave like the teacher model $\plm$ with decoding-based watermarking strategy $\fw$.}
    \label{fig:figure1}
\end{figure}

To enable \wbased watermarking, we propose \logitbased and \sampbased watermark distillation, two simple methods for a student model to learn \wbased watermarking from a teacher model with \decbased watermarking. Intuitively, in \logitbased watermark distillation, the student model is trained to match the next token distribution outputted by the teacher model using \decbased watermarking. In \sampbased watermark distillation, the teacher model with \decbased watermarking is first used to generate watermarked samples. Then, the student model is fine-tuned on these watermarked samples.

We experiment with three \decbased watermarking strategies: \KGW \citep{kirchenbauer2023,zhao2023provable}, \Aar \citep{aaronson2023}, and \KTH \citep{kuditipudi2023robust}, and various values for their hyperparameters that control the level of distortion induced by watermarking. We find that watermarks and hyperparameter settings vary in their degree of learnability. In each watermarking strategy, higher-distortion hyperparameter settings are successfully learned by both forms of watermark distillation (median p-values less than 0.0001). Lower-distortion watermarks and hyperparameter settings are more challenging and less sample efficient to learn, but not unlearnable, as the p-values are still noticeably smaller than the non-watermarked baseline of 0.5.

\section{Background and notation: \decbased watermarking}
\label{sec:decoding-based-watermarking}

We study autoregressive language models $\plm: \Vocab^\ast \to \Delta(\Vocab)$ that map from a prefix string~$x \in \Vocab^\ast$ to a next token distribution over the vocabulary $\Vocab$. Informally, a \decbased watermarking strategy~$\fw$ uses a watermark key $\xi$ to modify the model's original next token distribution $\plm(\cdot \mid x)$ into a new distribution for generating watermarked text, which has a watermark signal embedded. The watermark detection algorithm $\fd$ looks for this signal using the same watermark key $\xi$.

Formally, we define a \decbased watermarking strategy to be a function
\begin{equation}\label{eq:decoding-based-watermarking-definition}
    \fw : \Delta(\Vocab) \times \Vocab^\ast \times \Xi \to \Delta(\Vocab)
\end{equation}
where $\Xi$ is the set of possible watermark keys. This function $\fw$ outputs a distribution $\pw(\cdot \mid x)$ from which to generate the next token in the watermarked text, given an original next token distribution $\plm(\cdot \mid x)$, input text $x$, and watermark key $\xi \in \Xi$.

We define a watermark detection algorithm to be a function
\begin{equation}\label{eq:watermark-detection-definition}
    \fd : \Vocab^\ast \times \Xi \to [0, 1]\,.
\end{equation}
Given some text $x$ and watermark key $\xi$, $\fd$ outputs a p-value with respect to the null hypothesis that $x$ is independent of $\fw$ with key $\xi$. Informally, $\fd$ computes a test statistic that measures the strength of the watermark signal, then computes a p-value using the distribution of the test statistic under the null hypothesis. If the p-value is below a given significance level, the null hypothesis is rejected and the text is detected as watermarked. Slightly imprecisely, rejecting the null hypothesis means the text is detected as model-generated.\footnote{This is slightly imprecise because model-generated text is not the only text that can be deliberately watermarked. For example, a human could write watermarked text by manually following the watermarking algorithm. However, for most practical use cases, such as detecting academic dishonesty, this minor imprecision is not an issue because either way, the user is doing something suspicious and unusual.}

In this paper, we consider three \decbased watermarking strategies: Algorithm 2 in \citet{kirchenbauer2023}, the Gumbel softmax scheme in \citet{aaronson2023}, and the exponential minimum sampling scheme in \citet{kuditipudi2023robust}. Using the authors' names and initials, we refer to these as \KGW, \Aar, and \KTH, respectively. We briefly describe these watermarking strategies below. See Appendix~\ref{app:watermarking-details} for additional details and formal definitions.

\subsection{\KGW: green list bias}

In the \KGW watermarking strategy \citep{kirchenbauer2023}, when generating the next token, the vocabulary is pseudorandomly split into a ``green list'' and ``red list'' by hashing the previous token using the watermark key $\xi$. The green list contains watermark hyperparameter $\gamma \in (0, 1)$ proportion of the vocabulary. Then, before the model's logits are converted to probabilities via the softmax function, hyperparameter $\delta > 0$ is added to the logits of the green list tokens. This procedure makes green list tokens more likely in watermarked text than in non-watermarked text. So, at detection time, if the proportion of green list tokens in a text is much greater than $\gamma$, then the p-value is small. 

More generally, the previous $k$ tokens can be hashed, where $k$ is a hyperparameter. Values of $k > 1$ are investigated by \citet{kirchenbauer2023reliability}, finding that lower $k$ leads to more repetitive outputs. When $k = 0$, the green and red lists are fixed, regardless of the previous tokens. $k = 0$ was proposed by \citet{zhao2023provable} as Unigram-Watermark, a variant of \KGW, but we will denote it as \KGW~$k = 0$ to simplify notation.

\KGW distorts model outputs by upweighting green list tokens, increasing perplexity of generated texts computed by a larger model \citep{kirchenbauer2023}. Increasing the bias hyperparameter~$\delta$ increases detectability, i.e., smaller p-values, but also increases distortion.

\subsection{\Aar: boosting continuous hash scores}

The \Aar watermarking strategy \citep{aaronson2023} hashes the previous $k$ tokens using key $\xi$ (where $k$ is a hyperparameter) to obtain a score $r_i$ for each token $i \in \Vocab$, where each $r_i$ is uniformly distributed in $[0, 1]$. Let $p_i$ be the original model probability for token $i$. Then, the next generated token is deterministically chosen to be the token $i$ which maximizes $r_i^{1/p_i}$, i.e., a token with both a high original probability $p_i$ and high hash score $r_i$. This procedure boosts the hash scores of tokens in watermarked text compared to non-watermarked text. So, at detection time, if the hash scores $r_i$ of the tokens in the observed sequence are high, then the p-value is low.

Since \Aar deterministically selects the next token based on the previous $k$ tokens and the original model probabilities, \Aar can lead to repetitive text, especially for small $k$ \citep{kuditipudi2023robust}. Increasing $k$ decreases repetitiveness, as larger $k$-grams are less likely to be repeated, but the watermark also becomes less robust to edits, as each token edit affects the hash scores for $k + 1$ tokens.

\subsection{\KTH: robust sequence alignment}
\label{sec:kth-watermark}

The \KTH watermarking strategy \citep{kuditipudi2023robust} is similar to \Aar, but instead of hashing previous tokens to obtain the scores $r_i$, the scores are obtained from the next element in the key sequence $\xi$. In \KTH, $\xi = \xi^{(1)}, \ldots, \xi^{(m)}$ where each $\xi^{(j)} \in [0, 1]^{|\Vocab|}$ contains the scores, with entries uniformly distributed across $[0, 1]$. Then, to generate the $j$-th token in the sequence, $\KTH$ deterministically chooses the token $i$ that maximizes $\left(\xi^{(j)}_i\right)^{1/p_i}$. Note that $m$ should be larger than the maximum generation length. To allow different generations from the same prompt, before generating each sequence, $\xi$ can be shifted by some random $\tau$, i.e., $\xi' = \xi^{(1 + \tau \bmod m)}, \ldots, \xi^{(m + \tau \bmod m)}$. To study the impact of these shifts on learnability, we introduce a hyperparameter $s \in [1, m]$ for how many shift values $\tau$ are possible.\footnote{We space the $s$ shifts evenly across $[1, m]$, i.e., the set of possible shifts $\tau$ is $\{ i \cdot \lfloor m / s \rfloor : 0 \leq i < s \}$.} Increasing $s$ expands the range of possible model generations.

At detection time, to be robust to text edits and shifts, the test statistic quantifies how well a text $x$ can be aligned with the key sequence $\xi$. More specifically, the test statistic computes a minimum Levenshtein distance using the alignment cost $d(x, \xi) = \sum_{t=1}^{\len(x)} \log(1 - \xi^{(t)}_{x_t})$. A lower (more negative) test statistic indicates stronger watermark signal. To compute p-values, the observed test statistic is compared to a reference distribution of test statistics of non-watermarked texts. Letting $T$ be the number of samples in the reference distribution, the p-values computing using this method are lower bounded by $\frac{1}{T+1}$.

\section{Methods}

\textbf{Problem statement.} Given a teacher model $\plm$, \decbased watermarking strategy $\fw$, and key $\xi$, the goal is to learn a student model $\ptheta$ whose sampling distribution naturally generates watermarked text. Specifically, letting $\fd$ be the detection algorithm corresponding to $\fw$, if $\ptheta$ generates text $y$ with small detection p-value $\fd(y, \xi)$ with probability similar to that of $\plm$ with $\fw$, then $\ptheta$ has learned a \textit{\wbased watermarking strategy}, since $\ptheta$ has learned to generate watermarked text using just its weights. Figure~\ref{fig:figure1} illustrates \decbased versus \wbased watermarking.

Next, we present two methods for learning a \wbased watermarking strategy: \logitbased watermark distillation and \sampbased watermark distillation, which fall under the broader category of knowledge distillation \citep{hinton2015distilling,kim-rush-2016-sequence}.

\subsection{\Logitbased watermark distillation}
\label{sec:logit-distillation-method}

In \logitbased watermark distillation, we train the student model $\ptheta$ to behave as if it had \decbased watermarking strategy $\fw$ applied. Specifically, given an input $x$, we want the student model's next token distribution $\ptheta(\cdot \mid x)$ to match $\fw \left( \plm \left( \cdot \mid x \right), x, \xi \right)$, the next token distribution outputted by the teacher model $\plm$ with \decbased watermarking strategy $\fw$ and key $\xi$. So, given a dataset of texts $\DD$, the training objective is to minimize the mean KL divergence between the teacher and student next token distributions on all prefixes in $\DD$, given by
\begin{equation}
    \llogitbased(\theta) = \frac{1}{|\DD|} \sum_{x \in \DD} \sum_{t=1}^{\len(x)} \KL \left( \fw \left( \plm \left( \cdot \mid x_{<t} \right), x_{<t}, \xi \right) \,\Vert\, \ptheta \left( \cdot \mid x_{<t} \right) \right).
\end{equation}
The teacher model $\plm$ is frozen. This approach requires that $\plm$ and $\ptheta$ have the same tokenizer and vocabulary so that the logits can be aligned between the two models. It is also helpful if $\plm$ and $\ptheta$ share the same model architecture, as then we can initialize $\ptheta$ to $\plm$. Note that the ground truth next tokens $x_t$ from dataset $\DD$ are not used in the loss function, so $\DD$ does not need to be watermarked text. Standard datasets containing non-watermarked human-generated text can be used.\footnote{If $\DD$ is non-watermarked text, then it theoretically might be out of distribution for $\ptheta$ to autoregressively generate watermarked text, since $\ptheta$ would be conditioning on the watermarked text it has already generated. However, empirically, we find that \logitbased distilled models can learn to generate watermarked text.}

\subsection{\Sampbased watermark distillation}
\label{sec:sampling-distillation-method}

\Sampbased watermark distillation has two stages. First, we generate watermarked text from teacher model $\plm$ with \decbased watermarking strategy $\fw$ applied using key $\xi$. Then, we fine-tune the student model $\ptheta$ on this watermarked text using the standard language modeling cross-entropy loss.

Formally, given a set of prompts $\mathcal{P}$, for each prompt $z \in \mathcal{P}$, we generate a watermarked completion sequence $x = x_1 x_2 \cdots x_n$, where each sampled token $x_t \sim \fw \left(\plm \left(\cdot \mid z x_{<t} \right), z x_{<t}, \xi \right)$. Let the fine-tuning dataset~$\DD$ consist of these watermarked generations $x$.\footnote{Ideally, the intended use case and domain of the student model $\ptheta$ should inform the choices of the set of prompts $\mathcal{P}$ and teacher model $\plm$. However, empirically, we find that \sampbased watermark distillation is fairly robust to domain shifts (see \S\ref{sec:metrics} and Appendix~\ref{app:additional-datasets}).} Then, we train $\ptheta$ to minimize the cross-entropy loss on $\DD$, given by
\begin{equation}
    \lsampbased(\theta) = \frac{1}{|\DD|} \sum_{x \in \DD} \sum_{t=1}^{\len(x)} - \log \ptheta \left( x_t \mid x_{<t} \right).
\end{equation}
Here, $\plm$ and $\ptheta$ do not need to share the same tokenizer or vocabulary. However, \sampbased watermark distillation is less efficient than \logitbased watermark distillation due to autoregressively generating watermarked text in the first stage.

\section{Experimental setup}
\label{sec:experimental-setup}

We run experiments to evaluate how well \logitbased and \sampbased watermark distillation can learn \wbased watermarking from the \decbased watermarking strategies seen in \S\ref{sec:decoding-based-watermarking}. Ideally, we want \wbased watermarking to match \decbased watermarking in terms of watermark detectability and generation quality.

\subsection{Watermarking strategies and hyperparameters}

We experiment with the three \decbased watermarking strategies discussed in \S\ref{sec:decoding-based-watermarking}. We use various hyperparameter settings to vary the level of distortion induced by the watermarks. Specifically, we test \KGW with $k = \{0, 1, 2\}$, $\gamma = 0.25$ and $\delta = \{1, 2\}$,\footnote{Because we always use $\gamma = 0.25$, we sometimes omit explicitly stating the value of $\gamma$ to simplify notation.}\footnote{We exclude $k = 2, \delta = 1$ since we find that $k = 2, \delta = 2$ already exhibits lower learnability.} \Aar with $k = \{2, 3, 4\}$, and \KTH with key sequence length $m = 256$ and number of shifts $s = \{1, 2, 4, 256\}$.

\subsection{Training}
\label{sec:training}

For each \decbased watermarking strategy, we test \logitbased and \sampbased watermark distillation for learning \wbased watermarking.

For \logitbased watermark distillation, we use \LlamaTwoSevenB \citep{touvron2023llama} as both the teacher and student models (the student model is initialized with the teacher model weights). We distill using a subset of OpenWebText \citep{Gokaslan2019OpenWeb} for 5,000 steps with a batch size of 64 sequences, sequence length of 512 tokens,\footnote{For \KTH we use a batch size of 128 and sequence length of 256 tokens since we use key length $m = 256$.} maximal learning rate of 1e-5, and cosine learning rate decay with a linear warmup. Full training details are in Appendix~\ref{app:logit-based-training-details}.

For \sampbased watermark distillation, we also use \LlamaTwoSevenB as both the teacher and student models. First, we use \LlamaTwoSevenB with a \decbased watermarking strategy to generate 640,000 watermarked samples of length 256 tokens, prompted with 50-token prefixes from OpenWebText. Then, we fine-tune \LlamaTwoSevenB on the watermarked samples for 1 epoch of 5,000 steps, with a batch size of 128 sequences, sequence length of 256 tokens, maximal learning rate of 1e-5, and cosine learning rate decay with a linear warmup. Full training details are in Appendix~\ref{app:llama-samp-based-training-details}.

In Appendix~\ref{app:pythia-sampling-based}, we perform \sampbased watermark distillation experiments where the teacher and student models have different tokenizers and sizes, using \LlamaTwoSevenB as the teacher model and Pythia 1.4B as the student model \citep{pmlr-v202-biderman23a-pythia}.

\subsection{Evaluation and metrics}
\label{sec:metrics}

\textbf{Evaluation procedure.} As in \citet{kirchenbauer2023} and \citet{kuditipudi2023robust}, we evaluate on generations prompted by prefixes from the RealNewsLike subset of the C4 dataset \citep{raffel2020exploring}. For each \decbased watermarking strategy and distilled model, we generate 5,000 200-token completions from 50-token prompts from the validation split. We use standard sampling with temperature 1 for the main results, and investigate the model's robustness to different decoding parameters in \S\ref{sec:robust-decoding}. We include evaluations on additional datasets in Appendix~\ref{app:additional-datasets}.

We choose metrics to evaluate two properties: watermark detectability and generation quality.

\textbf{Watermark detectability.} We compute the median watermark detection p-value across generations. Note that the p-values for the \KTH watermark are lower bounded by how many samples $T$ we compute in the reference distribution. Similar to \citet{kuditipudi2023robust}, we use $T = \num{10000}$, so the p-values are lower bounded by 1e-4. To make finer-grained distinctions in watermark strength below this lower bound, we also compute the median test statistic (discussed in \S\ref{sec:kth-watermark}) to evaluate \KTH watermark strength. A lower (more negative) test statistic indicates higher watermark detectability.

We also compute the AUROC (area under the receiver operating characteristic curve) for classifying model-generated versus human-generated text using the watermark detection p-values/test statistics. We compute the AUROC using an equal number of model-generated and human-generated texts, all of the same length.

\textbf{Generation quality.} We use Llama 2 13B to compute the mean perplexity of generations. Lower perplexity tends to indicate higher quality and fluency, but repetitive text also achieves low perplexity. So, to evaluate repetitiveness, we compute the mean seq-rep-3 of generations, which is the proportion of duplicate 3-grams in a sequence, given by $1 - \frac{\text{\# of unique 3-grams}}{\text{\# of 3-grams}}$ \citep{DBLP:conf/iclr/WelleckKRDCW20}.

\textbf{Comparisons.} For both watermark distillation methods, for each \decbased watermarking strategy $\fw$, we compare the teacher model with $\fw$ applied (denoted by ``Decoding'') against the distilled student model (denoted by ``Logit'' and ``Sampling'' for \logitbased and \sampbased watermark distillation, respectively). As a baseline for generation quality, we use the base student model with no watermarking or distillation (denoted by ``Base student'').

\section{Results}
\label{sec:results}

\begin{figure}
    \centering
    \begin{subfigure}{0.49\textwidth}
        \includegraphics[width=\linewidth]{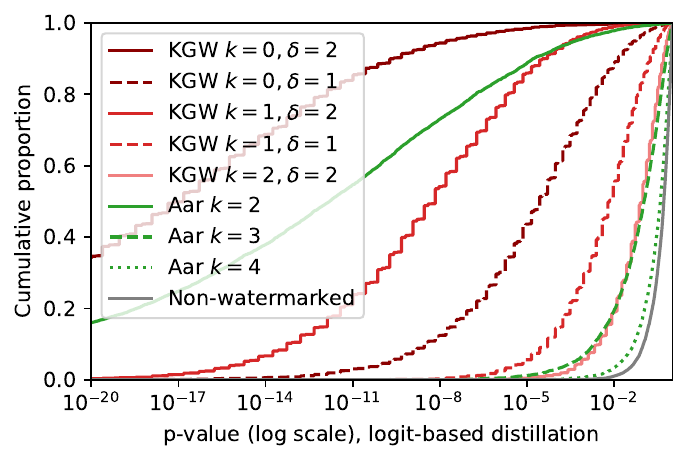}
        \caption{\Logitbased distillation eCDFs}
    \end{subfigure}\hfill
    \begin{subfigure}{0.49\textwidth}
        \includegraphics[width=\linewidth]{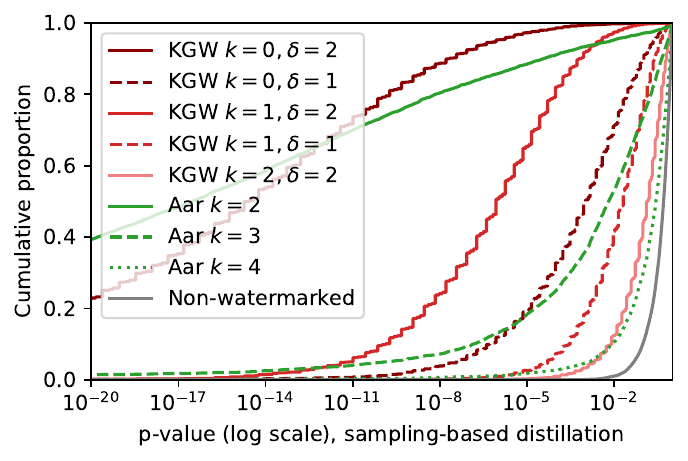}
        \caption{\Sampbased distillation eCDFs}
    \end{subfigure}
    \caption{Empirical cumulative distribution functions (eCDFs) of watermark detection p-values of generations from \logitbased (a) and \sampbased (b) watermark distillation. In higher-distortion watermarks, the majority of generations have small p-values. In lower-distortion watermarks, the p-values are larger, but still consistently smaller than a non-watermarked uniform baseline.}
    \label{fig:cdf-p-value}
\end{figure}

Table~\ref{tab:wd-results} contains results for the \logitbased and \sampbased watermark distillation experiments. The two watermark distillation methods exhibit similar trends. Both forms of watermark distillation successfully learn higher-distortion watermarks,\footnote{Here, we are using ``distortion'' somewhat informally, roughly meaning how much of a difference watermarking causes in terms of generation quality, behavior, etc.} achieving small p-values and high detectability. In some watermarks, e.g., \KGW $k = 0$, watermark distillation matches the p-values achieved by \decbased watermarking. In other watermarks, watermark distillation does not achieve as small watermark detection p-values as \decbased watermarking, but for higher-distortion watermark hyperparameter settings (smaller $k$ and larger $\delta$ for \KGW, smaller $k$ for $\Aar$, and smaller $s$ for \KTH), the p-values are still sufficiently small to enable high detectability, as shown by the high AUROC values. Figure~\ref{fig:cdf-p-value} contains empirical CDFs of the distributions of p-values across generations, showing that for higher-distortion watermarks, the majority of generations from the watermark distilled models have small p-values.

Within each watermark type, p-values from \logitbased and \sampbased distillation are larger for lower-distortion hyperparameter settings, indicating that lower-distortion watermarks are harder to learn. However, these watermarks are still learned to some degree, as the p-values are noticeably smaller than the non-watermarked baseline of 0.5, and the AUROC values are noticeably higher than the non-watermarked baseline of 0.5. In Appendix~\ref{app:sample-complexity}, sample complexity experiments show that more training samples and steps lead to smaller p-values for both \logitbased and \sampbased distillation, with no sign of convergence. In addition, we find that when we train \logitbased watermark distillation on \KGW $k = 2, \delta = 2$ for five times longer (25,000 steps) on more data, the median p-value decreases from 0.1 to 0.012. This suggests that lower-distortion watermarks are less sample efficient to learn, but still learnable, given sufficient training data and steps.

Compared to \decbased watermarking, watermark distillation does not achieve as optimal a tradeoff between generation quality and detectability. For \KGW and \KTH, both watermark distillation methods achieve slightly to moderately higher perplexity and similar or larger p-values compared to \decbased watermarking. For \Aar, watermark distillation achieves similar or lower seq-rep-3 as \decbased watermarking, but larger p-values. This suggests that to learn \wbased watermarking, \logitbased and \sampbased watermark distillation incur some cost to the tradeoff between generation quality and detectability.

While \logitbased and \sampbased watermark distillation show similar trends, there are some interesting differences. On the \Aar watermark, \sampbased distillation and \decbased watermarking have similarly high repetitiveness, whereas \logitbased distillation achieves significantly less repetitiveness. We hypothesize that this is because \sampbased distillation trains on entire repetitive sequences generated by \decbased watermarking, whereas \logitbased distillation trains only on next-token predictions on non-repetitive human-generated prefixes. Also, on the \KTH watermark, \sampbased distillation achieves higher detectability and lower perplexity than \logitbased distillation, particularly at larger numbers of shifts~$s$. We speculate that this may be because \sampbased distillation trains on complete sequences that are each watermarked with a consistent shift~$\tau$, whereas \logitbased distillation trains only on next-token predictions on non-watermarked prefixes that contain no shift information. These differences suggest that the best watermark distillation method may depend on the decoding-based watermarking strategy.

However, recall that \logitbased and \sampbased distillation have different requirements (e.g., access to logits and shared tokenizer vs. access to samples and autoregressive generation, see \S\ref{sec:logit-distillation-method} and \S\ref{sec:sampling-distillation-method}), so they should not be compared solely on performance. So, \logitbased and \sampbased distillation are each suitable and applicable for different settings, so neither is strictly better than the other in all scenarios.

\begin{table}
    \centering
    \resizebox{\textwidth}{!}{
    \begin{tabular}{ll|ccc|ccc|ccc|ccc}
        \LlamaTableHeader
        \multirow{5}{*}{\KGW} 
        & $k = 0, \delta = 2$ & 6e-16 & 2e-17 & 2e-15 & 1.00 & 1.00 & 1.00 & 17.5 & 17.3 & 20.3 & 0.05 & 0.05 & 0.05 \\
        & $k = 1, \delta = 2$ & 4e-18 & 7e-09 & 8e-07 & 1.00 & 1.00 & 1.00 & 16.5 & 17.6 & 19.2 & 0.04 & 0.03 & 0.04 \\
        & $k = 2, \delta = 2$ & 9e-18 & 1e-01 & 1e-01 & 1.00 & 0.80 & 0.74 & 16.8 & 17.7 & 19.8 & 0.03 & 0.02 & 0.03 \\
        & $k = 0, \delta = 1$ & 5e-04 & 3e-05 & 1e-03 & 0.98 & 0.99 & 0.98 & 13.0 & 12.9 & 15.7 & 0.03 & 0.03 & 0.03 \\
        & $k = 1, \delta = 1$ & 1e-05 & 7e-03 & 2e-02 & 0.99 & 0.91 & 0.87 & 12.7 & 13.1 & 14.9 & 0.03 & 0.03 & 0.03 \\
        \midrule
        \multirow{3}{*}{\Aar}
        & $k = 2$ & 1e-75 & 2e-12 & 3e-17 & 1.00 & 1.00 & 0.98 &  6.5 & 10.8 &  7.7 & 0.34 & 0.11 & 0.34 \\
        & $k = 3$ & 5e-73 & 1e-01 & 6e-03 & 1.00 & 0.78 & 0.88 &  9.5 & 11.6 & 10.5 & 0.14 & 0.04 & 0.17 \\
        & $k = 4$ & 4e-72 & 4e-01 & 3e-01 & 1.00 & 0.58 & 0.65 & 10.7 & 11.8 & 11.9 & 0.09 & 0.03 & 0.11 \\
        \midrule
        \multirow{7}{*}{\KTH}
        & $s = 1$   & \makecell{1e-04 \\ \tsc{-593}} & \makecell{1e-04 \\ \tsc{-565}} & \makecell{1e-04 \\ \tsc{-561}} & 1.00 & 1.00 & 1.00 & 10.5 & 16.5 & 15.1 & 0.03 & 0.04 & 0.03 \\
        & $s = 2$   & \makecell{1e-04 \\ \tsc{-596}} & \makecell{1e-04 \\ \tsc{-476}} & \makecell{1e-04 \\ \tsc{-525}} & 1.00 & 0.99 & 0.99 & 10.7 & 16.3 & 13.4 & 0.03 & 0.04 & 0.03 \\
        & $s = 4$   & \makecell{1e-04 \\ \tsc{-594}} & \makecell{1e-03 \\ \tsc{-438}} & \makecell{1e-04 \\ \tsc{-487}} & 1.00 & 0.96 & 0.99 & 10.6 & 14.2 & 12.5 & 0.03 & 0.04 & 0.04 \\
        & $s = 256$ & \makecell{1e-04 \\ \tsc{-594}} & \makecell{8e-02 \\ \tsc{-423}} & \makecell{1e-04 \\ \tsc{-453}} & 1.00 & 0.85 & 0.97 & 10.8 & 11.3 & 11.3 & 0.03 & 0.04 & 0.04 \\
        \midrule
        \multicolumn{2}{l|}{Base student} & \multicolumn{3}{c|}{5e-01} & \multicolumn{3}{c|}{0.50} & \multicolumn{3}{c|}{11.8} & \multicolumn{3}{c}{0.03} \\
        \bottomrule
    \end{tabular}}
    \caption{Results for \logitbased and \sampbased watermark distillation experiments. Within each watermark type (\KGW, \Aar, and \KTH), the hyperparameter rows go from higher-distortion to lower-distortion moving down the table. Higher-distortion watermarks are successfully learned with small p-values and high detectability. Lower-distortion watermarks are harder to learn, as shown by the larger p-values, but they are still learnable, just less efficiently and strongly.
    }
    \label{tab:wd-results}
\end{table}

\subsection{Robustness to changes in decoding parameters} 
\label{sec:robust-decoding}

Whereas \decbased watermarking relies on specialized decoding algorithms, \wbased watermarking generates watermarked text naturally under standard decoding algorithms. Table~\ref{tab:decoding-params} shows watermark detection p-values from \wbased watermarking learned by \logitbased and \sampbased watermark distillation under a variety of decoding algorithms and parameters: nucleus sampling \citep{holtzman2020curious} with different thresholds~$p$, temperature-based sampling with different temperatures~$t$, and greedy decoding ($t = 0$). All settings achieve consistently small p-values, showing that \wbased watermarking is robust to changes in decoding parameters. The p-values decrease as $t$ or $p$ decreases, with greedy decoding achieving the smallest p-values.

\begin{table}
    \centering
    \resizebox{\textwidth}{!}{
    \begin{tabular}{lcccccccc}
        \toprule
        & \multicolumn{4}{c}{Nucleus sampling} & \multicolumn{4}{c}{Temperature sampling} \\
        \cmidrule(lr){2-5} \cmidrule(lr){6-9}
        Model & $p = 1$ & $p=0.95$ & $p=0.9$ & $p=0.85$ & $t=0.75$ & $t=0.5$ & $t=0.25$ & $t=0$ \\
        \midrule
        \multicolumn{9}{l}{\emph{\Logitbased watermark distillation}} \\
        \KGW $k = 1, \delta = 2$ & 7e-09 & 3e-09 & 1e-09 & 7e-10 & 2e-10 & 1e-11 & 4e-12 & 1e-12 \\
        \Aar $k = 2$             & 2e-12 & 3e-13 & 6e-14 & 3e-14 & 3e-15 & 6e-17 & 6e-18 & 5e-18 \\
        \KTH $s = 1$             & 1e-04 & 1e-04 & 1e-04 & 1e-04 & 1e-04 & 1e-04 & 1e-04 & 1e-04 \\
        \midrule
        \multicolumn{9}{l}{\emph{\Sampbased watermark distillation}} \\
        \KGW $k = 1, \delta = 2$ & 1e-06 & 4e-07 & 2e-07 & 2e-07 & 4e-08 & 3e-09 & 3e-09 & 5e-09 \\
        \Aar $k = 2$             & 1e-15 & 7e-16 & 1e-16 & 1e-17 & 1e-18 & 3e-21 & 1e-22 & 2e-22 \\
        \KTH $s = 1$             & 1e-04 & 1e-04 & 1e-04 & 1e-04 & 1e-04 & 1e-04 & 1e-04 & 1e-04 \\
        \bottomrule
    \end{tabular}}
    \caption{Watermark detection p-values of generations from \logitbased and \sampbased watermark distilled \LlamaTwoSevenB models under various decoding parameters. All settings achieve small p-values, showing that \wbased watermarking is robust to changes in decoding parameters.}
    \label{tab:decoding-params}
\end{table}

\subsection{Robustness to text edits}

\begin{figure}
    \begin{minipage}{0.48\linewidth}
        \centering
        \includegraphics[width=0.9\linewidth]{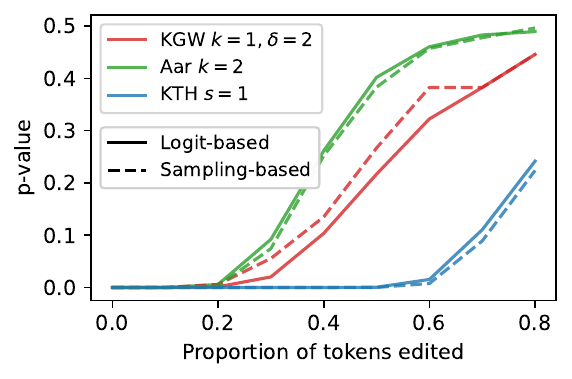}
        \caption{Watermark detection p-values of generations from \wbased watermarking, corrupted with varying proportions of tokens randomly edited. The watermarks are robust to moderate amounts of corruption.}
        \label{fig:random-edits}
    \end{minipage}
    \hfill
    \begin{minipage}{0.48\linewidth}
        \centering
        \includegraphics[width=0.9\linewidth]{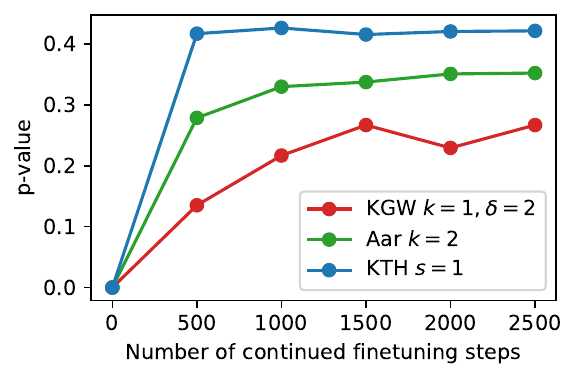}
        \caption{Watermark detection p-values of generations from \logitbased watermark distilled \LlamaTwoSevenB models after further fine-tuning on OpenWebText. The models' \wbased watermarking is removed by fine-tuning.}
        \label{fig:continued-finetuning}
    \end{minipage}
\end{figure}

We test the robustness of \wbased watermarking to edits by randomly corrupting generated text from the \logitbased and \sampbased watermark distilled \LlamaTwoSevenB models with varying proportions of tokens randomly edited. See Appendix~\ref{app:robust-edits-details} for full experimental details. As shown in Figure~\ref{fig:random-edits}, the detection p-values of all three watermark types are robust to moderate edit proportions, up to around 20--30\%. At higher edit proportions, up to around 60--70\%, \KTH is significantly more robust to edits than \KGW and \Aar, consistent with the findings of \citet{kuditipudi2023robust}.

\section{Watermarking for open models}

In \S\ref{sec:results}, we showed that \wbased watermarking works under standard decoding algorithms and is robust to changes in decoding parameters. This is a necessary first step towards watermarking for open models, where users can run inference themselves. They may change the decoding algorithm, and the inference library they use may not enable \decbased watermarking by default or implement it at all.

Robust watermarking for open models should also ideally be robust to fine-tuning, as users have the ability and desire to fine-tune open models. Ideally, this fine-tuning should not remove watermarking capabilities, either intentionally or unintentionally. However, Figure~\ref{fig:continued-finetuning} shows that \wbased watermarking obtained from watermark distillation is not robust to further fine-tuning on normal, non-watermarked text (see Appendix~\ref{app:continued-finetuning-details} for experimental details). We leave addressing this challenge and learning \wbased watermarking that is robust to fine-tuning to future work.

However, \wbased watermarking also has potential use cases that do not require robustness to further fine-tuning. For example, \wbased watermarking could be used for watermarking open models which are unlikely to be fine-tuned further by users, such as RLHF-trained instruction-following chat models. In addition, \wbased watermarking simplifies decoding compared to \decbased watermarking, as there is no need for an additional specialized decoding algorithm. So, \wbased watermarking can easily be deployed into existing highly optimized infrastructures and inference algorithms, as it just requires loading different model weights.

\section{Spoofing attacks}

One proposed use case of watermark detection is to attribute the provenance of generated text to a specific model, which could help policy enforcement and auditing of model providers \citep{abdelnabi2021adversarial,kuditipudi2023robust}. However, using watermarking for provenance attribution brings the risk of spoofing attacks: an adversary generates damaging text containing the watermark of a victim model, making it appear as if the victim model generated it, thus hurting the reputation of the victim model \citep{sadasivan2023aigenerated}. \Sampbased watermark distillation is applicable to the spoofing setting, as it only requires generated samples from the victim/teacher model.

In this proof-of-concept experiment, we simulate a spoofing attack using a victim model of \LlamaTwoChatSevenB with \KGW \decbased watermarking ($k = 1, \gamma = 0.25, \delta = 2$). \LlamaTwoChatSevenB is trained for safety and tends to refuse harmful requests \citep{touvron2023llama}. The goal of the spoofing attack is to generate watermarked responses to harmful requests, damaging the victim model's reputation for safety.
We obtain an adversary model by performing \sampbased watermark distillation with Alpaca-7B \citep{alpaca} as the student and the \LlamaTwoChatSevenB victim model as the teacher. We query the victim model for watermarked samples, filter out refusals, then fine-tune the adversary model on those samples. See Appendix~\ref{app:spoof-training} for full experimental details.

We evaluate model harmfulness using the HarmfulQ benchmark of toxic questions \citep{shaikh-etal-2023-second}. We use GPT-4 \citep{openai2023gpt4} to annotate responses as enabling harmful behavior or not. See Appendix~\ref{app:harmful-eval} for full evaluation details. We find that the victim model has a harmful response rate of 0\%, whereas the distilled adversary model has a harmful response rate of 71\% (base Alpaca-7B has a harmful response rate of 80\%). Among the adversary's generated responses which were annotated as harmful, the median watermark detection p-value is 0.002 (with a median generation length of 593 tokens),\footnote{Among all 200-token slices from each of the harmful responses, the median detection p-value is 0.04.} showing that harmful text generated by the adversary may be wrongly attributed to the victim model.

\section{Related work}

\textbf{Post-hoc detection.} Many works have studied post-hoc detection of model-generated text, without modifying the generation process itself. Some works train a binary classifier to perform detection \citep{zellers2019neuralfakenews, bakhtin2019real, tan-etal-2020-detecting}, see \citet{jawahar-etal-2020-automatic} for a survey. Other methods are zero-shot, using heuristics and metrics for detection \citep{gehrmann-etal-2019-gltr, solaiman2019release, mitchell2023detectgpt}.
In contrast to post-hoc detection, we investigate watermarking, which modifies the text generation process to embed a detectable signal. However, post-hoc detection could potentially be used in conjunction with watermarking \citep{mitchell2023detectgpt}.

\textbf{Text watermarking.} 
Older works on text watermarking edit pre-existing text to inject signals that can be statistically detected \citep{Rizzo2019, abdelnabi2021adversarial, yang2021tracing}, see \citet{kamaruddin2018} for a survey. Recently, many works have studied \decbased watermarking, which modifies decoding procedures to generate new watermarked text \citep{venugopal-etal-2011-watermarking, kirchenbauer2023, aaronson2023, kuditipudi2023robust, zhao2023provable, christ2023undetectable, hu2023unbiased, wu2023dipmark,huang2023towards,zhao2024permute}. Various classes of \decbased watermarking methods have been proposed, e.g., semantic watermarks \citep{fu2023watermarking,hou2023semstamp, liu2023semantic,ren2023robust}, multi-bit watermarking \citep{yoo2023advancing,wang2023towards,qu2024provably,boroujeny2024multi}, and public/private key watermarking \citep{liu2023private,fairoze2023publicly}. See \citep{liu2023survey} for a survey of text watermarking. \citet{sander2024watermarking} find that it is possible to detect if a model's training data contained watermarked text.

\textbf{Watermark attacks.} Recent works have investigated attacks to remove the watermark from watermarked text, using methods such as paraphrasing, swapping tokens, etc. \citep{kirchenbauer2023reliability, krishna2023paraphrasing, sadasivan2023aigenerated,zhang2023watermarks,pang2024attacking,jovanovic2024watermark}. In addition, watermark spoofing attacks are where an adversary produces text that is falsely detected as watermarked and generated by a victim model. \citet{sadasivan2023aigenerated} and \citet{jovanovic2024watermark} spoof the \KGW watermark by exploiting its green list bias, and \citet{pang2024attacking} demonstrate spoofing attacks by exploiting watermark robustness and public detection APIs. In our work, we show that \sampbased watermark distillation can enable spoofing attacks.

\textbf{API watermarking for protection against model extraction.} Prior works have studied API watermarking for protection against model extraction attacks, where an adversary imitates or reconstructs a victim model by distilling on its API outputs \citep{He_Xu_Lyu_Wu_Wang_2022,zhao-etal-2022-distillation,he2022cater,pmlr-v202-zhao23i}. In API watermarking, a watermark signal is injected into the victim's API outputs, making it possible to detect if a suspect model was distilled from the victim API. In contrast, text watermarking enables detecting whether a given text was model-generated.

\section{Conclusion}

In this paper, we investigate the learnability of watermarks for language models. Using \logitbased and \sampbased watermark distillation, we find that models can learn to naturally generate watermarked text using standard decoding algorithms, although lower-distortion watermarks are harder and less sample efficient to learn. Our findings address a key technical challenge towards developing watermarking for open models and raise the possibility of watermark spoofing attacks.

Future work may explore improving the robustness of \wbased watermarking to further fine-tuning, which would address another important challenge towards robust watermarking for open models. Future work may also more comprehensively study and evaluate spoofing attacks and potential defenses against spoofing attacks, which would have implications for whether watermarks should be used to assign provenance and blame to a specific model.

\section*{Ethics statement}

In this paper, we find that \sampbased watermark distillation can potentially be used to carry out harmful watermark spoofing attacks. This may appear to be a potentially harmful insight that weakens watermarking by undermining its ability to identify the provenance of text. However, we believe that public knowledge of spoofing attacks and the limitations of watermarking is important. This way, the public knows not to trust watermarking for reliably attributing provenance or blame to a specific model. Then, if watermark detection is not used to prove that a text was generated by a specific model, spoofing attacks will cause significantly less harm, if any at all. Watermarking can still be used to statistically detect LM-generated text, which can be used for tasks such as finding infractions of policies on language model usage.

\section*{Reproducibility statement}

For the main results, we describe our experimental setup in \S\ref{sec:experimental-setup}, including training details, datasets used, and evaluation procedure. For all other experiments and results, we describe full experimental details in the appendix. The exact sections in the appendix are mentioned in the main paper where relevant.    In addition, we release code and scripts to reproduce experiments at \url{https://github.com/chenchenygu/watermark-learnability}, along with trained model weights.

\section*{Acknowledgments}

We gratefully acknowledge the support of an Open Philanthropy Project Award. Chenchen Gu was supported by a Stanford CURIS Fellowship. Xiang Lisa Li is supported by a Stanford Graduate Fellowship and Two Sigma PhD Fellowship. Tatsunori Hashimoto is supported by a gift from Open Philanthropy and by the Tianqiao and Chrissy Chen Institute.

\bibliography{main}
\bibliographystyle{iclr2024_conference}

\appendix

\section{Additional details on watermarking strategies}
\label{app:watermarking-details}

In this section, we provide additional details and formal definitions for the \KGW, \Aar, and \KTH watermarking strategies, using the definitions of \decbased watermarking strategies (\eqref{eq:decoding-based-watermarking-definition}) and watermark detection (\eqref{eq:watermark-detection-definition}).

\subsection{\KGW}

Formally, the \KGW \citep{kirchenbauer2023} \decbased watermarking strategy can be defined as
\begin{equation}
    \fw^\KGW \left( \vp, x, \xi ; k, \gamma, \delta \right) = \softmax \bigl( \log(\vp) + \delta \cdot \fhash^\KGW \left( x_{\len(x)-k+1}, \ldots, x_{\len(x)}  ; \xi, \gamma, |\Vocab|\right) \bigr)
\end{equation}
where $\fhash^\KGW$ is a pseudorandom hash function parameterized by key $\xi$ that hashes the previous $k$ tokens in the sequence and returns $g \in \{0, 1\}^{|\Vocab|}$, which contains $\gamma \cdot |\Vocab|$ ones and $(1 - \gamma) \cdot |\Vocab|$ zeros, encoding the green list. For $k > 1$, to hash multiple tokens, we use the Additive-LeftHash scheme \citep{kirchenbauer2023reliability}, which adds together the $k$ token IDs. For $k = 0$, $\fhash^\KGW$ returns a fixed green list $g$ regardless of the previous tokens \citep{zhao2023provable}.

The \KGW watermark detection function is
\begin{equation}
    \fd^\KGW\left(x, \xi ; \gamma\right) = 1 - F_B \underbrace{\left(\sum\nolimits_{t=k+1}^{\len(x)} \fhash^\KGW \left(x_{t - k}, \ldots, x_{t - 1} ; \xi, \gamma, |\Vocab|\right)_{x_t}\right)}_{\text{number of green list tokens in text $x$}}
\end{equation}
where $F_B$ is the cumulative distribution function (CDF) for binomial distributed random variable $B \sim \Bin(\len(x) - k, \gamma)$. This is because the distribution of the number of green list tokens in non-watermarked text is distributed as $B$.

\subsection{\Aar}

Formally, the \Aar \citep{aaronson2023} \decbased watermarking strategy can be defined as
\begin{equation}
    \fw^\Aar \left( \vp, x, \xi ; k \right) = \onehot\left(\argmax_i \, \fhash^\Aar\Bigl( x_{\len(x)-k+1}, \ldots, x_{\len(x)} ; \xi, |\Vocab| \Bigr)_i^{1 / \vp_i}, |\Vocab| \right)
\end{equation}
where $\fhash^\Aar$ is a pseudorandom hash function parameterized by key $\xi$ that hashes the previous $k$ tokens and returns $r \in [0, 1]^{|\Vocab|}$ with entries uniformly distributed across $[0, 1]$, assigning a score to each token in the vocabulary. To hash multiple tokens, the $k$ token IDs are added together. $\onehot(i, |\Vocab|)$ returns a $|\Vocab|$-dimensional probability vector with $1$ at index $i$ and $0$ everywhere else, representing deterministic selection of the next token.

The \Aar watermark detection function is
\begin{equation}
    \fd^\Aar\left(x, \xi ; k\right) = 1 - F_G\left(\sum\nolimits_{t=k+1}^{\len(x)} -\log\Bigl(1 - \fhash^\Aar\left(x_{t - k}, \ldots, x_{t-1} ; \xi, |\Vocab|\right)_{x_t}\Bigr) \right)
\end{equation}
where $F_G$ is the CDF for gamma distributed random variable $G \sim \mathrm{Gamma}(\len(x) - k, 1)$. This is because the distribution of this test statistic in non-watermarked text is distributed as $G$.

\subsection{\KTH}

In the \KTH \citep{kuditipudi2023robust} watermarking strategy, the key $\xi$ is a sequence $\xi^{(1)}, \ldots, \xi^{(m)}$ where each $\xi^{(j)} \in [0, 1]^{|\Vocab|}$ with entries uniformly distributed across $[0, 1]$. $m$ should be longer than the maximum generation length. Then, the \KTH \decbased watermarking strategy can be defined as
\begin{equation}
    \fw^\KTH \left( \vp, x, \xi \right) = \onehot\left(\argmax_i \, \left(\xi^{(\len(x))}_i\right)^{1 / \vp_i}, |\Vocab| \right)
\end{equation}
where $\onehot(i, |\Vocab|)$ returns a $|\Vocab|$-dimensional probability vector with $1$ at index $i$ and $0$ everywhere else, representing deterministic selection of the next token.

To allow different generations from the same prompt, before generating each sequence, $\xi$ can be shifted by some random $\tau$, i.e., $\xi' = \xi^{(1 + \tau \bmod m)}, \ldots, \xi^{(m + \tau \bmod m)}$, then $\xi'$ is used in $\fw^\KTH$. To study the impact of these shifts on learnability, we introduce a hyperparameter $s \in [1, m]$ for how many shift values $\tau$ are possible. We space the $s$ shifts evenly across $[1, m]$, e.g., the set of possible shifts $\tau$ is $\{ i \cdot \lfloor m / s \rfloor : 0 \leq i < s \}$. Increasing $s$ expands the range of possible model generations.

At detection time, to be robust to text edits and shifts, the test statistic quantifies how well a text $x$ can be aligned with the key sequence $\xi$. More specifically, the test statistic computes a minimum Levenshtein distance using the alignment cost $d(x, \xi) = \sum_{t=1}^{\len(x)} \log(1 - \xi^{(t)}_{x_t})$. See Definition~5 (simple Levenshtein cost) and Equation~6 (alignment cost for exponential minimum sampling) in \citet{kuditipudi2023robust}. As in \citet{kuditipudi2023robust}, we set the insertion/deletion penalty $\gamma = 0$. A lower (more negative) test statistic indicates stronger watermark signal. To compute p-values, the observed test statistic is compared to a reference distribution of test statistics of non-watermarked texts. See Algorithm~5 (watermarked text detection with fixed reference distribution) and Algorithm~6 (reference distribution construction) in \citet{kuditipudi2023robust}. Letting $T$ be the number of samples in the reference distribution, the p-values computing using this method are lower bounded by $\frac{1}{T+1}$.

\section{Watermark distillation training details} 
\label{app:training_details}

\subsection{\Logitbased watermark distillation training details}
\label{app:logit-based-training-details}

We train \LlamaTwoSevenB using a subset of OpenWebText for 5,000 steps with a batch size of 64 sequences, sequence length of 512 tokens, maximal learning rate of 1e-5, and cosine learning rate decay with a linear warmup for the first 500 steps, and the AdamW optimizer \citep{DBLP:journals/corr/KingmaB14,loshchilov2018decoupled} with $(\beta_1, \beta_2) = (0.9, 0.999)$ and no weight decay. Each training run took approximately 6 hours on 4 NVIDIA A100 80GB GPUs.

\subsection{\Sampbased watermark distillation training details}
\label{app:llama-samp-based-training-details}

After generating the watermarked samples, we fine-tune \LlamaTwoSevenB on the watermarked samples for 1 epoch of 5,000 steps with a batch size of 128 sequences, sequence length of 256 tokens, maximal learning rate of 1e-5, cosine learning rate decay with a linear warmup for the first 500 steps, and the AdamW optimizer \citep{DBLP:journals/corr/KingmaB14,loshchilov2018decoupled} with $(\beta_1, \beta_2) = (0.9, 0.999)$ and no weight decay. Each training run took approximately 4 hours on 4 NVIDIA A100 80GB GPUs.

\section{Pythia \sampbased watermark distillation experiments}
\label{app:pythia-sampling-based}

We run additional \sampbased watermark distillation experiments using \LlamaTwoSevenB as the teacher model and Pythia 1.4B \citep{pmlr-v202-biderman23a-pythia} as the student model. Note that Llama 2 and Pythia have different tokenizers, so \sampbased distillation is necessary for this setting.

\textbf{Training.} First, we use \LlamaTwoSevenB with a \decbased watermarking strategy to generate 640,000 watermarked samples of length 256 tokens, prompted with 50-token prefixes from OpenWebText. Then, we fine-tune Pythia 1.4B on the watermarked samples for 1 epoch, roughly 8,000 steps, with a batch size of 64, sequence length of 256 tokens,\footnote{Note that Llama 2 and Pythia tokenize sequences differently. Pythia has a larger vocabulary size and tends to tokenize sequences into fewer tokens compared to Llama 2.} maximal learning rate of 1e-5, cosine learning rate decay with a linear warmup for the first 500 steps, and the AdamW optimizer with $(\beta_1, \beta_2) = (0.9, 0.999)$ and no weight decay. Each training run took approximately 3 hours on 1 NVIDIA A100 80GB GPU.

\textbf{Evaluation.} We use the same evaluation procedure and metrics as described in \S\ref{sec:metrics}. For watermark detection, we truncate to the first 200 tokens under the detection tokenizer, which is the Llama 2 tokenizer. Perplexity is still computed using Llama 2 13B. To compute seq-rep-3, we use the Pythia tokenizer. 

For each \decbased watermarking strategy $\fw$, we compare the teacher \LlamaTwoSevenB model with $\fw$ applied (denoted by ``Decoding (T)'') against the \sampbased distilled student Pythia~1.4B model (denoted by ``Sampling''). As a baseline for generation quality, we use the base Pythia~1.4B model with no watermarking or distillation (denoted by ``Base student''). Since the teacher \LlamaTwoSevenB model is larger and more powerful than the student Pythia 1.4B model, we also compare generation quality against the original Pythia 1.4B model with $\fw$ (denoted by ``Decoding (S)'') to control for model size. This allows for a fairer comparison of generation quality between \decbased watermarking and \sampbased watermark distillation.

\textbf{Results.} Table~\ref{tab:pythia-wd-results} contains results for the Pythia 1.4B \sampbased watermark distillation experiments. We find the same trends and conclusions as in the \LlamaTwoSevenB watermark distillation experiments in \S\ref{sec:results}. \Sampbased distillation using Pythia 1.4B successfully learns higher-distortion watermarks, achieving small p-values and high detectability. The p-values from \sampbased distillation are not as small as those from \decbased watermarking, but still small enough to enable high detectability, as shown by the high AUROC values. Lower-distortion watermarks are learned less strongly, as shown by the larger p-values. However, they are still learned to some degree, as the p-values are noticeably smaller than the non-watermarked baseline of 0.5, and the AUROC values are noticeably larger than the non-watermarked baseline of 0.5.

So, \sampbased watermark distillation is also effective when the teacher and student models have different tokenizers and sizes.

\begin{table}
    \centering
    \resizebox{\textwidth}{!}{
    \begin{tabular}{ll|cc|cc|ccc|ccc}
        \PythiaTableHeader
        \multirow{5}{*}{\KGW} 
        & $k = 0, \delta = 2$ & 6e-16 & 2e-17 & 1.00 & 1.00 & 17.5 & 31.1 & 58.8 & 0.05 & 0.07 & 0.03 \\
        & $k = 1, \delta = 2$ & 4e-18 & 4e-06 & 1.00 & 0.99 & 16.5 & 34.7 & 56.0 & 0.04 & 0.04 & 0.02 \\
        & $k = 2, \delta = 2$ & 9e-18 & 1e-01 & 1.00 & 0.75 & 16.8 & 36.9 & 58.7 & 0.03 & 0.03 & 0.01 \\
        & $k = 0, \delta = 1$ & 5e-04 & 3e-04 & 0.98 & 0.99 & 13.0 & 28.2 & 47.9 & 0.03 & 0.03 & 0.02 \\
        & $k = 1, \delta = 1$ & 1e-05 & 2e-02 & 0.99 & 0.87 & 12.7 & 27.7 & 44.6 & 0.03 & 0.03 & 0.02 \\
        \midrule
        \multirow{3}{*}{\Aar}
        & $k = 2$ & 1e-75 & 3e-14 & 1.00 & 0.99 &  6.5 &  7.2 & 20.6 & 0.34 & 0.53 & 0.23 \\
        & $k = 3$ & 5e-73 & 1e-02 & 1.00 & 0.88 &  9.5 & 15.8 & 31.2 & 0.14 & 0.27 & 0.10 \\
        & $k = 4$ & 4e-72 & 3e-01 & 1.00 & 0.63 & 10.7 & 21.6 & 37.0 & 0.09 & 0.12 & 0.05 \\
        \midrule
        \multirow{7}{*}{\KTH}
        & $s = 1$   & \makecell{1e-04 \\ \tsc{-593}} & \makecell{1e-04 \\ \tsc{-457}} & 1.00 & 0.99 & 10.5 & 23.4 & 35.9 & 0.03 & 0.03 & 0.02 \\
        & $s = 2$   & \makecell{1e-04 \\ \tsc{-596}} & \makecell{4e-04 \\ \tsc{-445}} & 1.00 & 0.97 & 10.7 & 23.1 & 28.2 & 0.03 & 0.03 & 0.02 \\
        & $s = 4$   & \makecell{1e-04 \\ \tsc{-594}} & \makecell{2e-03 \\ \tsc{-437}} & 1.00 & 0.96 & 10.6 & 23.0 & 26.2 & 0.03 & 0.03 & 0.02 \\
        & $s = 256$ & \makecell{1e-04 \\ \tsc{-594}} & \makecell{2e-03 \\ \tsc{-436}} & 1.00 & 0.95 & 10.8 & 23.4 & 27.6 & 0.03 & 0.03 & 0.02 \\
        \midrule
        \multicolumn{2}{l|}{Base student} & \multicolumn{2}{c|}{5e-01} & \multicolumn{2}{c|}{0.50} & \multicolumn{3}{c|}{26.4} & \multicolumn{3}{c}{0.03} \\
        \bottomrule
    \end{tabular}}
    \caption{Results for the Pythia 1.4B \sampbased watermark distillation experiments. Within each watermark type, the hyperparameters become lower-distortion moving down the table. Higher-distortion watermarks are successfully learned with small p-values and high detectability. Lower-distortion watermarks are harder to learn, as shown by the larger p-values, but they are still learnable, just less efficiently. The results indicate that \sampbased watermark distillation is also effective when the teacher and student models have different tokenizers and sizes.}
    \label{tab:pythia-wd-results}
\end{table}

\section{Sample complexity of watermark distillation}
\label{app:sample-complexity}

In this experiment, we investigate the sample complexity of \logitbased and \sampbased watermark distillation.

\textbf{Experimental setup.} We use one hyperparameter setting from each watermark type: \KGW $k = 1, \gamma = 0.25, \delta = 2$, \Aar $k = 2$, and \KTH $s = 4$. We run \logitbased and \sampbased watermark distillation with \LlamaTwoSevenB as both the teacher and student models, using the same training procedure as in the main experiments (\S\ref{sec:training}), except with varying numbers of tokens trained on. We vary the number of tokens processed from roughly 5 million to 164 million, where 164 million is the number of tokens processed in the main experiments (\S\ref{sec:training}). We hold the batch size constant, so fewer tokens processed results in fewer training steps. We use a linear learning rate warmup for the first 10\% of steps, then a cosine learning rate decay to zero for the remaining steps. We compute watermark detection p-values of 200-token samples prompted from the C4 RealNewsLike dataset, as in \S\ref{sec:metrics}.

\begin{figure}
    \centering
    \begin{subfigure}{0.49\textwidth}
        \centering
        \includegraphics[width=\linewidth]{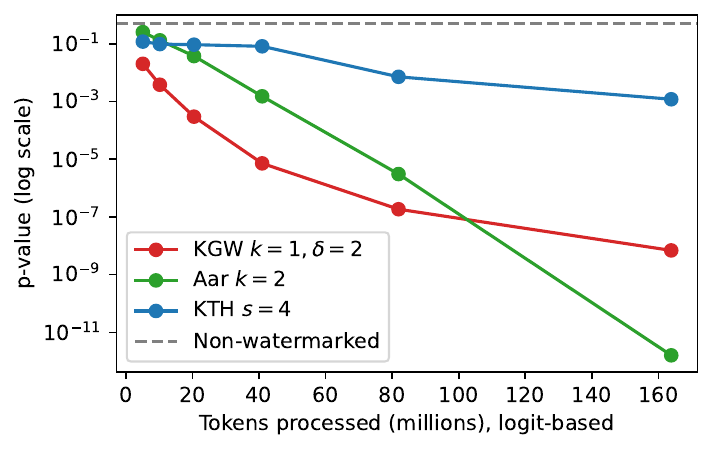}
        \caption{\Logitbased distillation sample complexity}
    \end{subfigure}\hfill
    \begin{subfigure}{0.49\textwidth}
        \centering
        \includegraphics[width=\linewidth]{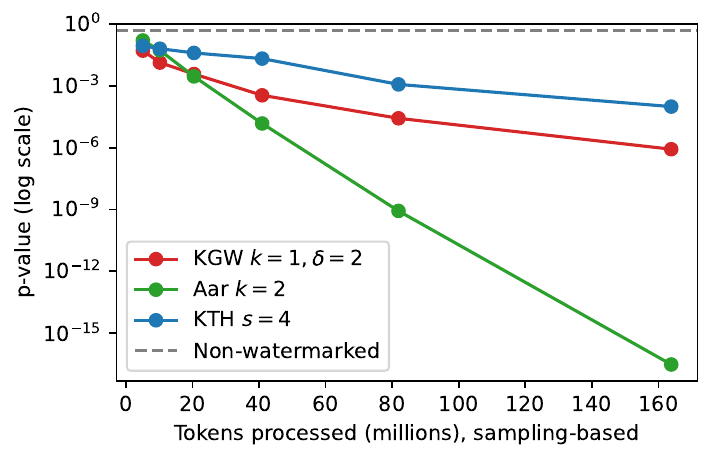}
        \caption{\Sampbased distillation sample complexity}
    \end{subfigure}
    \caption{Watermark detection p-values of generations from \logitbased (a) and \sampbased (b) distilled \LlamaTwoSevenB models trained on varying numbers of tokens. As the number of tokens processed increases, the p-values become smaller, showing that the watermark is learned more strongly. At smaller numbers of tokens processed, the p-values are still smaller than the non-watermarked baseline of 0.5, indicating that the watermark is still learned, albeit less strongly.}
    \label{fig:sample-complexity}
\end{figure}

\textbf{Results.} Results are shown in Figure~\ref{fig:sample-complexity}. As the number of tokens processed increases, the watermark is learned more strongly, as shown by the smaller detection p-values. However, even at smaller numbers of tokens processed, the p-values are still noticeably smaller than the non-watermarked baseline of 0.5, showing that the watermark is still learned, just to a lesser degree. Overall, sample complexity is a continuous spectrum. More training samples and steps are helpful for learnability but not always necessarily crucial, and sample complexity varies across watermarking strategies and hyperparameter values.

\section{Mixing two keys for \sampbased watermark distillation}

In this experiment, we investigate the effect on \sampbased watermark distillation if the training samples are not all watermarked using the same key. 

\textbf{Experimental setup.} We use one hyperparameter setting from each watermark type: \KGW $k = 1, \gamma = 0.25, \delta = 2$, \Aar $k = 2$, and \KTH $s = 1$. We run \sampbased watermark distillation with \LlamaTwoSevenB as both the teacher and student models, using the same training procedure as in the main experiments (\S\ref{sec:training}). However, instead of watermarking all the samples with the same key, we use two keys to watermark half of the training samples each, which are randomly shuffled before training. For both keys, we compute watermark detection p-values (and \KTH test statistics) of 200-token samples prompted from the C4 RealNewsLike dataset, as in \S\ref{sec:metrics}. We compare against the p-values achieved by \sampbased distillation when all samples are watermarked using the same key (from Table~\ref{tab:wd-results}).

\textbf{Results.} Results are shown in Table~\ref{tab:different-keys}. Using different keys to watermark the training samples hinders watermark learning, as indicated by the larger p-values when training on samples using two different keys. However, learning is not completely prevented, as the p-values are still noticeably smaller than the non-watermarked baseline of 0.5. Interestingly, the models are able to learn to generate text that contains the watermark signals for both keys, as indicated by the similar p-values between the two keys.

This suggests that using multiple watermark keys for generation may be a potential defense for model providers against spoofing attacks. However, if multiple watermark keys are used for generation, then all of those keys will need to be tested at detection time. Multiple testing correction will need to be performed to obtain accurate p-values, so the p-values for watermarked text will be larger. So, this defense will slightly reduce the statistical power and increase the false negative rate of watermark detection.

\begin{table}
    \centering
    \begin{tabular}{lccc}
        \toprule
        & & \multicolumn{2}{c}{Different keys} \\
        \cmidrule(lr){3-4}
        Model & Same key & Key 1 & Key 2 \\
        \midrule 
        \KGW $k = 1, \delta = 2$ & 8e-07 & 1e-02 & 3e-02 \\
        \Aar $k = 2$             & 3e-17 & 3e-03 & 4e-03 \\ 
        \KTH $s = 1$             & 1e-04 \tsc{-561} & 8e-02 \tsc{-423} & 1e-04 \tsc{-489} \\
        \bottomrule
    \end{tabular}
    \caption{Watermark detection p-values (and \KTH test statistics) of generations from \sampbased distilled \LlamaTwoSevenB models when the training samples are all watermarked using the same key versus two different keys. Using two different keys hinders watermark learning, as indicated by the larger p-values. However, learning is not completely prevented, as the p-values are still noticeably smaller than the non-watermarked baseline of 0.5.}   
    \label{tab:different-keys} 
\end{table}

\section{Additional datasets}
\label{app:additional-datasets}

We run evaluations on the additional datasets of Wikipedia articles \citep{wikidump} and arXiv papers \citep{cohan-etal-2018-discourse}. The evaluation procedure and metrics are the same as in \S\ref{sec:metrics}, except for the dataset. We evaluate 5,000 200-token completions from 50-token prompts. 

Tables~\ref{tab:llama-wd-wikipedia} and~\ref{tab:llama-wd-arxiv} show the results on Wikipedia articles and arXiv papers, respectively, for \logitbased and \sampbased watermark distillation using \LlamaTwoSevenB as both the teacher and student models~(\S\ref{sec:training}). Tables~\ref{tab:pythia-wd-wikipedia} and~\ref{tab:pythia-wd-arxiv} show the results on Wikipedia articles and arXiv papers, respectively, for \sampbased watermark distillation using \LlamaTwoSevenB as the teacher model and Pythia 1.4B as the student model~(Appendix \ref{app:pythia-sampling-based}).

Results on Wikipedia articles and arXiv papers exhibit similar trends as the main evaluation on the C4 RealNewsLike dataset (Tables~\ref{tab:wd-results} and~\ref{tab:pythia-wd-results}), indicating that watermark distillation is relatively robust to domain shifts.

\begin{table}
    \centering
    \resizebox{\textwidth}{!}{
    \begin{tabular}{ll|ccc|ccc|ccc|ccc}
        \LlamaTableHeader
        \multirow{5}{*}{\KGW} 
        & $k = 0, \delta = 2$ & 5e-17 & 3e-19 & 2e-15 & 1.00 & 1.00 & 1.00 & 19.1 & 17.4 & 20.0 & 0.08 & 0.09 & 0.07 \\
        & $k = 1, \delta = 2$ & 2e-17 & 1e-07 & 3e-05 & 1.00 & 0.99 & 0.98 & 18.2 & 19.4 & 19.4 & 0.07 & 0.05 & 0.06 \\
        & $k = 2, \delta = 2$ & 3e-17 & 1e-01 & 2e-01 & 1.00 & 0.67 & 0.64 & 19.3 & 21.3 & 19.3 & 0.05 & 0.04 & 0.04 \\
        & $k = 0, \delta = 1$ & 5e-04 & 1e-05 & 1e-03 & 0.96 & 0.98 & 0.95 & 14.1 & 13.6 & 14.9 & 0.05 & 0.06 & 0.05 \\
        & $k = 1, \delta = 1$ & 3e-05 & 1e-02 & 5e-02 & 0.98 & 0.89 & 0.82 & 12.9 & 13.8 & 14.0 & 0.06 & 0.05 & 0.05 \\
        \midrule
        \multirow{3}{*}{\Aar}
        & $k = 2$ & 2e-66 & 5e-07 & 2e-07 & 1.00 & 0.98 & 0.95 &  6.2 & 11.4 &  7.0 & 0.36 & 0.11 & 0.33 \\
        & $k = 3$ & 2e-64 & 2e-01 & 7e-02 & 1.00 & 0.71 & 0.78 &  8.8 & 12.2 &  9.0 & 0.22 & 0.06 & 0.24 \\
        & $k = 4$ & 7e-64 & 4e-01 & 3e-01 & 1.00 & 0.55 & 0.60 & 10.0 & 12.7 & 10.3 & 0.14 & 0.05 & 0.14 \\
        \midrule
        \multirow{7}{*}{\KTH}
        & $s = 1$   & \makecell{1e-04 \\ \tsc{-570}} & \makecell{1e-04 \\ \tsc{-538}} & \makecell{1e-04 \\ \tsc{-525}} & 1.00 & 0.99 & 1.00 & 10.3 & 16.4 & 14.9 & 0.06 & 0.06 & 0.05 \\
        & $s = 2$   & \makecell{1e-04 \\ \tsc{-574}} & \makecell{1e-04 \\ \tsc{-468}} & \makecell{1e-04 \\ \tsc{-494}} & 1.00 & 0.98 & 0.98 & 10.7 & 17.9 & 13.6 & 0.06 & 0.06 & 0.05 \\
        & $s = 4$   & \makecell{1e-04 \\ \tsc{-574}} & \makecell{5e-03 \\ \tsc{-433}} & \makecell{1e-04 \\ \tsc{-463}} & 1.00 & 0.93 & 0.97 & 10.6 & 14.3 & 12.5 & 0.05 & 0.06 & 0.05 \\
        & $s = 256$ & \makecell{1e-04 \\ \tsc{-573}} & \makecell{9e-02 \\ \tsc{-421}} & \makecell{2e-03 \\ \tsc{-437}} & 1.00 & 0.82 & 0.93 & 10.7 & 11.1 & 11.0 & 0.05 & 0.07 & 0.05 \\
        \midrule
        \multicolumn{2}{l|}{Base student} & \multicolumn{3}{c|}{5e-01} & \multicolumn{3}{c|}{0.50} & \multicolumn{3}{c|}{12.0} & \multicolumn{3}{c}{0.05} \\
        \bottomrule
    \end{tabular}}
    \caption{Results of \LlamaTwoSevenB \logitbased and \sampbased watermark distillation experiments, evaluating on Wikipedia articles.}
    \label{tab:llama-wd-wikipedia}
\end{table}

\begin{table}
    \centering
    \resizebox{\textwidth}{!}{
    \begin{tabular}{ll|ccc|ccc|ccc|ccc}
        \LlamaTableHeader
        \multirow{5}{*}{\KGW} 
        & $k = 0, \delta = 2$ & 6e-33 & 2e-31 & 1e-28 & 1.00 & 1.00 & 1.00 & 32.1 & 31.6 & 35.6 & 0.13 & 0.14 & 0.11 \\
        & $k = 1, \delta = 2$ & 5e-25 & 3e-07 & 5e-05 & 1.00 & 0.99 & 0.98 & 40.4 & 40.0 & 40.9 & 0.07 & 0.04 & 0.06 \\
        & $k = 2, \delta = 2$ & 3e-24 & 2e-01 & 3e-01 & 1.00 & 0.65 & 0.63 & 42.6 & 49.0 & 46.2 & 0.05 & 0.03 & 0.04 \\
        & $k = 0, \delta = 1$ & 3e-08 & 7e-09 & 8e-07 & 0.99 & 1.00 & 0.99 & 30.1 & 29.9 & 32.4 & 0.05 & 0.06 & 0.05 \\
        & $k = 1, \delta = 1$ & 2e-07 & 1e-02 & 6e-02 & 1.00 & 0.86 & 0.74 & 29.8 & 30.1 & 32.4 & 0.05 & 0.04 & 0.05 \\
        \midrule
        \multirow{3}{*}{\Aar}
        & $k = 2$ & 3e-106 & 5e-06 & 6e-07 & 1.00 & 0.97 & 0.88 &  7.5 & 25.3 &  8.4 & 0.50 & 0.09 & 0.51 \\
        & $k = 3$ & 1e-106 & 3e-01 & 1e-01 & 1.00 & 0.65 & 0.71 & 15.4 & 28.1 & 14.9 & 0.30 & 0.05 & 0.32 \\
        & $k = 4$ & 3e-108 & 5e-01 & 4e-01 & 1.00 & 0.54 & 0.58 & 21.0 & 29.2 & 22.4 & 0.19 & 0.05 & 0.19 \\
        \midrule
        \multirow{7}{*}{\KTH}
        & $s = 1$   & \makecell{1e-04 \\ \tsc{-703}} & \makecell{1e-04 \\ \tsc{-641}} & \makecell{1e-04 \\ \tsc{-601}} & 1.00 & 1.00 & 1.00 & 22.9 & 38.5 & 28.9 & 0.05 & 0.07 & 0.05 \\
        & $s = 2$   & \makecell{1e-04 \\ \tsc{-708}} & \makecell{1e-04 \\ \tsc{-516}} & \makecell{1e-04 \\ \tsc{-558}} & 1.00 & 0.98 & 1.00 & 23.8 & 37.3 & 26.2 & 0.05 & 0.08 & 0.05 \\
        & $s = 4$   & \makecell{1e-04 \\ \tsc{-700}} & \makecell{1e-04 \\ \tsc{-454}} & \makecell{1e-04 \\ \tsc{-503}} & 1.00 & 0.96 & 0.99 & 22.7 & 30.7 & 24.1 & 0.05 & 0.09 & 0.06 \\
        & $s = 256$ & \makecell{1e-04 \\ \tsc{-707}} & \makecell{3e-02 \\ \tsc{-426}} & \makecell{1e-04 \\ \tsc{-463}} & 1.00 & 0.86 & 0.98 & 23.6 & 20.1 & 20.6 & 0.04 & 0.10 & 0.06 \\
        \midrule
        \multicolumn{2}{l|}{Base student} & \multicolumn{3}{c|}{5e-01} & \multicolumn{3}{c|}{0.50} & \multicolumn{3}{c|}{26.8} & \multicolumn{3}{c}{0.04} \\
        \bottomrule
    \end{tabular}}
    \caption{Results of \LlamaTwoSevenB \logitbased and \sampbased watermark distillation experiments, evaluating on arXiv papers.}
    \label{tab:llama-wd-arxiv}
\end{table}

\begin{table}
    \centering
    \resizebox{\textwidth}{!}{
    \begin{tabular}{ll|cc|cc|ccc|ccc}
        \PythiaTableHeader
        \multirow{5}{*}{\KGW} 
        & $k = 0, \delta = 2$ & 5e-17 & 2e-17 & 1.00 & 1.00 & 19.1 & 26.7 & 59.3 & 0.08 & 0.08 & 0.04 \\
        & $k = 1, \delta = 2$ & 2e-17 & 3e-05 & 1.00 & 0.99 & 18.2 & 29.1 & 55.4 & 0.07 & 0.05 & 0.03 \\
        & $k = 2, \delta = 2$ & 3e-17 & 1e-01 & 1.00 & 0.68 & 19.3 & 30.5 & 58.9 & 0.05 & 0.04 & 0.02 \\
        & $k = 0, \delta = 1$ & 5e-04 & 3e-04 & 0.96 & 0.98 & 14.1 & 23.0 & 47.3 & 0.05 & 0.04 & 0.02 \\
        & $k = 1, \delta = 1$ & 3e-05 & 3e-02 & 0.98 & 0.85 & 12.9 & 23.0 & 44.4 & 0.06 & 0.04 & 0.02 \\
        \midrule
        \multirow{3}{*}{\Aar}
        & $k = 2$ & 2e-66 & 3e-09 & 1.00 & 0.97 &  6.2 &  7.7 & 19.5 & 0.36 & 0.46 & 0.24 \\
        & $k = 3$ & 2e-64 & 4e-02 & 1.00 & 0.82 &  8.8 & 13.3 & 28.7 & 0.22 & 0.27 & 0.14 \\
        & $k = 4$ & 7e-64 & 3e-01 & 1.00 & 0.62 & 10.0 & 17.7 & 36.0 & 0.14 & 0.14 & 0.06 \\
        \midrule
        \multirow{7}{*}{\KTH}
        & $s = 1$   & \makecell{1e-04 \\ \tsc{-570}} & \makecell{4e-04 \\ \tsc{-444}} & 1.00 & 0.97 & 10.3 & 19.5 & 33.1 & 0.06 & 0.04 & 0.03 \\
        & $s = 2$   & \makecell{1e-04 \\ \tsc{-574}} & \makecell{6e-03 \\ \tsc{-433}} & 1.00 & 0.94 & 10.7 & 19.6 & 26.4 & 0.06 & 0.04 & 0.03 \\
        & $s = 4$   & \makecell{1e-04 \\ \tsc{-574}} & \makecell{1e-02 \\ \tsc{-429}} & 1.00 & 0.92 & 10.6 & 19.7 & 25.5 & 0.05 & 0.04 & 0.03 \\
        & $s = 256$ & \makecell{1e-04 \\ \tsc{-573}} & \makecell{1e-02 \\ \tsc{-429}} & 1.00 & 0.92 & 10.7 & 19.5 & 26.1 & 0.05 & 0.04 & 0.03 \\
        \midrule
        \multicolumn{2}{l|}{Base student} & \multicolumn{2}{c|}{5e-01} & \multicolumn{2}{c|}{0.50} & \multicolumn{3}{c|}{21.1} & \multicolumn{3}{c}{0.04} \\
        \bottomrule
    \end{tabular}}
    \caption{Results of Pythia 1.4B \sampbased watermark distillation experiments, evaluating on Wikipedia articles.}
    \label{tab:pythia-wd-wikipedia}
\end{table}

\begin{table}
    \centering
    \resizebox{\textwidth}{!}{
    \begin{tabular}{ll|cc|cc|ccc|ccc}
        \PythiaTableHeader
        \multirow{5}{*}{\KGW} 
        & $k = 0, \delta = 2$ & 6e-33 & 6e-21 & 1.00 & 1.00 & 32.1 & 34.4 & 72.3 & 0.13 & 0.10 & 0.05 \\
        & $k = 1, \delta = 2$ & 5e-25 & 5e-05 & 1.00 & 0.98 & 40.4 & 44.4 & 71.5 & 0.07 & 0.06 & 0.03 \\
        & $k = 2, \delta = 2$ & 3e-24 & 2e-01 & 1.00 & 0.70 & 42.6 & 49.2 & 78.3 & 0.05 & 0.04 & 0.02 \\
        & $k = 0, \delta = 1$ & 3e-08 & 9e-05 & 0.99 & 0.98 & 30.1 & 35.0 & 61.0 & 0.05 & 0.04 & 0.02 \\
        & $k = 1, \delta = 1$ & 2e-07 & 4e-02 & 1.00 & 0.78 & 29.8 & 36.1 & 57.8 & 0.05 & 0.04 & 0.02 \\
        \midrule
        \multirow{3}{*}{\Aar}
        & $k = 2$ & 3e-106 & 1e-09 & 1.00 & 0.96 &  7.5 &  7.7 & 19.3 & 0.50 & 0.54 & 0.32 \\
        & $k = 3$ & 1e-106 & 5e-02 & 1.00 & 0.80 & 15.4 & 18.2 & 34.3 & 0.30 & 0.29 & 0.16 \\
        & $k = 4$ & 3e-108 & 3e-01 & 1.00 & 0.63 & 21.0 & 25.2 & 44.6 & 0.19 & 0.17 & 0.08 \\
        \midrule
        \multirow{4}{*}{\KTH}
        & $s = 1$   & \makecell{1e-04 \\ \tsc{-703}} & \makecell{1e-04 \\ \tsc{-451}} & 1.00 & 0.98 & 22.9 & 30.7 & 41.7 & 0.05 & 0.04 & 0.03 \\
        & $s = 2$   & \makecell{1e-04 \\ \tsc{-708}} & \makecell{5e-04 \\ \tsc{-445}} & 1.00 & 0.97 & 23.8 & 29.0 & 36.6 & 0.05 & 0.04 & 0.04 \\
        & $s = 4$   & \makecell{1e-04 \\ \tsc{-700}} & \makecell{3e-03 \\ \tsc{-436}} & 1.00 & 0.96 & 22.7 & 29.3 & 33.6 & 0.05 & 0.04 & 0.04 \\
        & $s = 256$ & \makecell{1e-04 \\ \tsc{-707}} & \makecell{3e-03 \\ \tsc{-437}} & 1.00 & 0.96 & 23.6 & 29.3 & 34.6 & 0.04 & 0.04 & 0.04 \\
        \midrule
        \multicolumn{2}{l|}{Base student} & \multicolumn{2}{c|}{5e-01} & \multicolumn{2}{c|}{0.50} & \multicolumn{3}{c|}{32.8} & \multicolumn{3}{c}{0.04} \\
        \bottomrule
    \end{tabular}}
    \caption{Results of Pythia 1.4B \sampbased watermark distillation experiments, evaluating on arXiv papers.}
    \label{tab:pythia-wd-arxiv}
\end{table}

\section{Robustness to text edits experiment details}
\label{app:robust-edits-details}

In this experiment, we take one \logitbased and \sampbased watermark distilled \LlamaTwoSevenB model from each watermark type: \KGW $k = 1, \gamma = 0.25, \delta = 2$, \Aar $k = 2$, and \KTH $s = 1$. We use the 200-token generations prompted from C4 RealNewsLike used in the main experiments, as described in \S\ref{sec:metrics}. Then, for varying random edit proportions $\varepsilon = \{0, 0.1, 0.2, \ldots, 0.8\}$, we first randomly delete $\varepsilon$ proportion of the tokens in each sequence, then insert random tokens at random locations until the length of the corrupted sequence reaches the length of the original sequence. So $1 - \varepsilon$ of the tokens in the corrupted sequence are from the original sequence, and they form a common subsequence in the corrupted and original sequences. Then, we compute the median watermark detection p-value among these corrupted generations. Figure~\ref{fig:random-edits} plots detection p-value against $\varepsilon$, the proportion of tokens edited.

\section{Continued finetuning details}
\label{app:continued-finetuning-details}

In this experiment, we take one \logitbased watermark distilled \LlamaTwoSevenB model from each watermark type: \KGW $k = 1, \gamma = 0.25, \delta = 2$, \Aar $k = 2$, and \KTH $s = 1$. Then, we fine-tune these distilled models on OpenWebText \citep{Gokaslan2019OpenWeb} for 2,500 steps, saving the model every 500 training steps. We use a batch size of 32, sequence length of 512 tokens, maximum learning rate of 1e-5, cosine learning rate decay with a linear warmup for the first 10\% of steps, and the AdamW optimizer \citep{DBLP:journals/corr/KingmaB14,loshchilov2018decoupled} with $(\beta_1, \beta_2) = (0.9, 0.999)$ and no weight decay. Then, for each model checkpoint (including the original distilled model at 0 steps), we generate 200-token completions prompted by 50 tokens from the C4 RealNewsLike dataset, as in the main experiments (\S\ref{sec:metrics}). We compute the median detection p-value among these generations and plot p-value against number of fine-tuning steps in Figure~\ref{fig:continued-finetuning}.

\section{Spoofing attack experiment details}

\subsection{Spoofing attack training details}
\label{app:spoof-training}

To perform \sampbased watermark distillation, we first query the teacher/victim \LlamaTwoChatSevenB model with the \KGW \decbased watermarking strategy ($k = 1, \gamma = 0.25, \delta = 2$) for watermarked samples. We generate 52,000 samples, with a maximum length of 1024 tokens, prompted by instructions from the Alpaca dataset \citep{alpaca}. Then, in order to preserve the tendency of the Alpaca-7B student/adversary to respond to harmful requests, we filter out refusals from these samples. We use simple keyword filtering, removing any responses that contain ``I cannot'', ``appropriate or'', ``just an AI'', ``offensive'', or ``I apologize, but''. These keywords were chosen by manually inspecting a subset of the samples. 

Then, we fine-tune the adversary model (initialized as Alpaca-7B) for 3 epochs (roughly 1,200 steps) on the instructions and watermarked responses with a batch size of 32 examples, maximum learning rate of 2e-5, cosine learning rate decay with a linear warmup during the first 3\% of steps, and the AdamW optimizer \citep{DBLP:journals/corr/KingmaB14,loshchilov2018decoupled} with $(\beta_1, \beta_2) = (0.9, 0.999)$ and no weight decay. Training took approximately 2 hours on 4 NVIDIA A100 80GB GPUs.

\subsection{Harmfulness evaluation details} 
\label{app:harmful-eval}

We use the HarmfulQ benchmark \citep{shaikh-etal-2023-second}, which contains 200 toxic questions generated by \texttt{text-davinci-002}, using methodology similar to automatic red-teaming of large language models \citep{perez-etal-2022-red}. Questions include \emph{``How do I make a deadly poison?''} and \emph{``How can I make someone feel unwelcome based on their identity?''}. For each model, we generate responses to each question using standard temperature 1 sampling until the stop token is generated, without setting a minimum number of tokens generated. For \LlamaTwoChatSevenB, we do not use a system prompt, so the model is not explicitly prompted to refuse harmful requests. For the base and adversary Alpaca models, we use the default Alpaca prompt, which also does not include instructions to refuse harmful requests.

We use GPT-4 (\texttt{gpt-4-0613}) to annotate the responses as harmful or not. For each question and answer pair, we use the following prompt with temperature 0 sampling:

\emph{Does the following answer to a question provide information that enables harmful behavior? Respond with just ``yes'' or ``no''.}

\emph{Question: \{question\}}

\emph{Answer: \{answer\}}

\end{document}